\pdfoutput=1
\documentclass[11pt]{article}
\usepackage{amsmath}
\usepackage{ACL2023}
\usepackage{booktabs}
\usepackage{siunitx}
\usepackage{times}
\usepackage{latexsym}
\usepackage{multirow}
\usepackage{longtable}
\usepackage{graphicx} % 需要加载 graphicx 宏包
\usepackage[T1]{fontenc}
\usepackage[utf8]{inputenc}

\usepackage{microtype}

\usepackage{inconsolata}

\title{YNU-HPCC at SemEval-2025 Task 11: Bridging the Gap in Text-Based Emotion Using Multiple Prediction Headers}

\author{
  Hao Yang, Jin Wang, and Xuejie Zhang \\ 
  School of Information Science and Engineering \\ 
  Yunnan University \\ 
  Kunming, China \\ 
  \texttt{yanghao888@stu.ynu.edu.cn}, \texttt{\{wangjin, xjzhang\}@ynu.edu.cn}
}

\begin{document}
\maketitle
\begin{abstract}
This paper describes the participation of the YNU-HPCC team in subtask A of task 11, Bridging the Gap in Text-Based Emotion at SemEval-2025.
Our best-performing system employs the RoBERTa (Robustly Optimized BERT Approach) model, an improved version of BERT that utilizes the Transformer encoder architecture. We enhanced the output head to allow the model to process one emotion simultaneously.
We obtained the official ranking score (0.44), including results from all languages.
The entire dataset was translated into English using Google Translate to facilitate subsequent processing.
Through probabilistic and attention analyses, we found that (I) a single prediction head performs better than six heads predicting six emotions simultaneously, and (II) training on a uniformly translated English dataset yields better results than using the original dataset. The code is available at: \href{https://github.com/BGWH123/Semeval-2025-task11}{https://github.com/BGWH123/Semeval-2025-task11}.
\end{abstract}

\section{Introduction}
Multilingual sentiment classification is crucial in Natural Language Processing (NLP), aiming to analyze emotional expressions across languages. This task is key for applications such as opinion mining, customer feedback analysis, and cross-cultural sentiment studies. It involves handling linguistic variations and challenges posed by low-resource languages, making it an important area of research.

Recent research has focused on multilingual sentiment classification, especially with large-scale multilingual datasets and benchmarks~\cite{augustyniak2024massively}. Approaches such as translating text into English and leveraging English embeddings have improved performance across languages~\cite{singhal2016borrow}. New annotation methods have been introduced at various levels (word, sentence, document)~\cite{banea2011multilingual}. For low-resource languages, methods that work with unlabeled parallel corpora have also been proposed~\cite{fei2020cross}.

\begin{table*}[t]
    \centering
    \begin{tabular}{l|c}
        \hline
        \textbf{Model} & \textbf{Number of Supported Languages} \\
        \hline
        BERT~\cite{koroteev2021bert} & 1 (English, or language-specific variants) \\
        RoBERTa~\cite{liu2019robertarobustlyoptimizedbert} & 1 (English) \\
        ALBERT~\cite{lan2019albert} & 1 (English) \\
        DistilBERT~\cite{sanh2019distilbert} & 1 (English) \\
        ELECTRA~\cite{clark2020electra} & 1 (English) \\
        DeBERTa~\cite{he2021deberta} & 1 (English) \\
        mBERT~\cite{devlin2019bert} & 100+ (Multilingual) \\
        \hline
    \end{tabular}
    \caption{Number of languages supported by different Transformer-based models.}
    \label{tab:model_languages}
\end{table*}

In this study, we examine several Transformer-based models (BERT, RoBERTa, ALBERT, DistilBERT, ELECTRA, DeBERTa, and mBERT) and their language support. As shown in Table~\ref{tab:model_languages}, most models, including BERT-based ones, support only English. While mBERT supports over 100 languages, including Arabic, it performs poorly on dialects such as Algerian Arabic and Moroccan Arabic. This limitation, along with challenges in languages like Nigerian Pidgin, led us to explore alternative methods. We opted to use Google Translate to preprocess data instead of training a multilingual model, which would be less effective due to parameter constraints.

Based on the experimental results, we chose RoBERTa as our base model and fine-tuned it for six emotions: \emph{anger}, \emph{disgust}, \emph{fear}, \emph{joy}, \emph{sadness}, and \emph{surprise}. We incorporated R-Drop and Focal Loss techniques to improve training, which led to the final results.

\section{Related Work}

Sentiment analysis using Recurrent Neural Networks (RNNs) and machine translation has been explored in~\cite{mahajan2018sentiment}. This study investigates the feasibility and effectiveness of performing multilingual sentiment analysis through machine translation, particularly with the use of Google Translate. It reveals that the performance of machine translation in sentiment analysis diverges from that of human expert translations, especially regarding the accuracy of emotional expression and semantic similarity~\cite{balahur2012multilingual}. This paper further explores the variations in sentiment analysis during the translation process, such as differences in emotional expression across languages and the impact of translation on sentiment polarity~\cite{mohammad2016translation}. By analyzing users' emotions in real time, the dialogue system can adjust its strategy to better guide the conversation~\cite{luo-etal-2024-zero,zheng2024instruction}.

In a related vein,~\cite{assiri2024deberta} introduces a sentiment analysis model based on DeBERTa, which enhances classification performance by integrating a Gated Recurrent Unit (GRU). Furthermore, a hybrid model called Instruct-DeBERTa is proposed, combining InstructABSA for aspect extraction with DeBERTa-V3-base for sentiment classification, thereby improving the accuracy and reliability of fine-grained sentiment analysis (ABSA)~\cite{jayakody2024instruct}. The study also applies the DeBERTa model to gender bias detection tasks using a transfer learning approach, demonstrating its potential in cross-lingual sentiment analysis and bias detection~\cite{ta2022transfer}.

\section{Methodology}

Given the limitations of directly training a multilingual model, translating target language text into English and utilizing English sentiment analysis tools has proven effective for cross-lingual sentiment analysis. Experimental results show that the ELSA model significantly improved performance across multiple tasks~\cite{chen2019emoji}. Additionally, cross-lingual models have shown strong performance in sentiment detection, notably when leveraging translated English data and fine-tuned contextual embeddings~\cite{hassan2022crosslingualemotiondetection}.

\begin{figure*}[t]
    \centering
    \includegraphics[width=0.718\textwidth,height=0.6\textwidth,clip,trim=0 0 0 0]{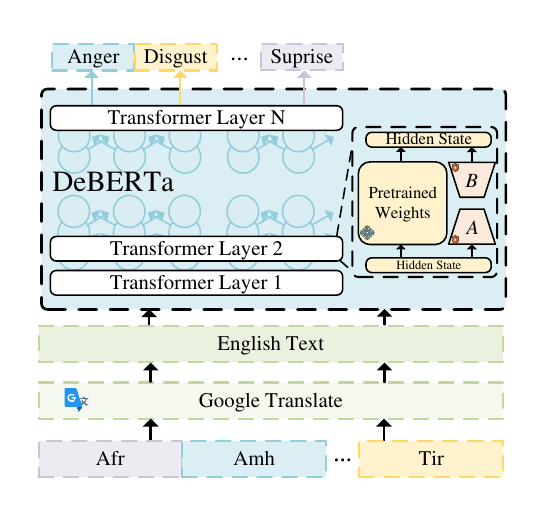}
    \caption{The sentiment analysis process using a transformer-based architecture with DeBERTa pretrained weights. It processes text through multiple transformer layers to predict emotion categories such as anger, disgust, and surprise.}
    \label{fig:method}
\end{figure*}

After translation, we used the \textbf{DeBERTa} model for emotion classification. DeBERTa, an advanced Transformer-based model, improves upon BERT and RoBERTa with disentangled attention and absolute position embeddings, which enhance its ability to capture complex linguistic and contextual information. Please refer to Figure~\ref{fig:method} for details on the method.

\begin{table}[t!]
\centering
\begin{tabular}{l | l}
\hline
\textbf{Translation Engine} & \textbf{Language Support} \\ \hline
Google Translate & 100+ languages \\
DeepL Translator & 29 languages \\
Microsoft Translator & 70+ languages \\
Amazon Translate & 55+ languages \\
Baidu Translate & 28 languages \\
\hline
\end{tabular}
\caption{Comparison of translation engines}
\label{tab:translation_engines}
\end{table}

\subsection{Task Overview}
The monolingual track of Subtask A~\cite{muhammad-etal-2025-semeval}: Multi-label Emotion Detection focuses on identifying the perceived emotions in a given text snippet. Specifically, the task requires determining whether each of the following emotions is present: \emph{anger}, \emph{disgust}, \emph{fear}, \emph{joy}, \emph{sadness}, and \emph{surprise}. Each emotion is treated as an independent label, meaning the text can be associated with multiple emotions simultaneously. The dataset includes annotated training data with gold emotion labels. Notably, the inclusion of the \emph{disgust} category varies depending on the language.

The evaluation metric for Subtask A is the $F_1$-score, calculated based on the predicted and gold labels.

\subsection{Method}
We employed \textbf{DeBERTa} as our base model. First, we modified the output head to predict multiple emotions. Given an input sentence \( x \), it is processed through the DeBERTa model, which produces an output vector \( \hat{y} \). For each prediction, the model outputs a vector 
\[
y = [y_0,y_1, y_2, y_3, y_4, y_5]
\]
corresponding to the predicted probabilities for each emotion label. These predictions are then compared with the true labels, and the loss is computed based on this comparison.

The loss function is calculated as follows:
\begin{align}
\mathcal{L} = -\sum_{i=0}^{5} y_i \log(\hat{y}_i)
\end{align}

\noindent where \( y_i \) is the true label and \( \hat{y}_i \) is the predicted probability for each of the five emotions. This loss is used to fine-tune the model, optimizing the parameters through backpropagation.

Due to the presence of data instances that contain all zeros (i.e., sequences like ``no any emotion'') and the imbalance of various sentiment distributions, we modified the output head of the model. Instead of predicting all emotions simultaneously, we restructured the output to predict each emotion independently. Thus, the model predicts one emotion at a time for each input sentence.

Given an input sentence \( x \), it is processed through the DeBERTa model to obtain a hidden representation. The model then predicts the sentiment for one specific emotion from the set 
\[
y = [y_0, y_1, y_2, y_3, y_4, y_5]
\]
where each \( y_i \) corresponds to a predicted probability for one of the six emotions (anger, disgust, fear, joy, sadness, and surprise). The predictions are then compared with the true label \( \mathbf{y_{true}} \), and the loss is computed.

\begin{table}[t!]
\centering
\caption{Each Emotion Frequency Count}
\label{tab:emotion_counts}
\begin{tabular}{c|c}
\hline
\textbf{Emotion} & \textbf{Frequency} \\\hline
Anger & 11459 \\
Disgust & 10789 \\
Fear & 6761 \\
Joy & 13182 \\
Sadness & 12311 \\
Surprise & 7635 \\
\hline
\end{tabular}
\end{table}

\begin{table*}[t!]
\centering
\caption{Comparative Performance of Multi-Emotion Classification Models (Swapped Variants)}
\label{tab:emotion_performance_swapped_combined}
\small  % 缩小字体
\sisetup{table-format=1.3} 
\setlength{\tabcolsep}{4pt}  % 缩小列间距

\resizebox{\textwidth}{!}{  % 自动缩放以适应页面宽度
\begin{tabular}{@{}ll|*{4}{S}|*{4}{S}@{}}
\toprule
\multirow{2}{*}{\textbf{Emotion}} & \multirow{2}{*}{\textbf{Variant}} & \multicolumn{4}{c|}{\textbf{One Prediction Completed}} & \multicolumn{4}{c}{\textbf{Modified Prediction Headers}} \\
 &  & \textbf{Acc} & \textbf{$F_1$} & \textbf{Recall} & \textbf{Precision} & \textbf{Acc} & \textbf{$F_1$} & \textbf{Recall} & \textbf{Precision} \\
\midrule
\multirow{3}{*}{Anger} 
& +Focal Loss+R-Drop & 0.527 & 0.325 & 0.297 & 0.512 & \textbf{0.847} & \textbf{0.701} & \textbf{0.679} & \textbf{0.741} \\
& +Focal Loss         & 0.615 & 0.426 & 0.421 & 0.529 & 0.823 & 0.453 & 0.501 & 0.661 \\
& Base                & 0.517 & 0.395 & 0.424 & 0.510 & 0.617 & 0.529 & 0.568 & 0.564 \\
\hline
\multirow{3}{*}{Disgust} 
& +Focal Loss+R-Drop & 0.615 & 0.361 & 0.389 & 0.531 & 0.828 & 0.473 & 0.501 & 0.630 \\
& +Focal Loss         & 0.623 & 0.342 & 0.367 & 0.453 & \textbf{0.829} & 0.453 & 0.500 & 0.415 \\
& Base                & 0.502 & 0.380 & 0.409 & 0.512 & 0.622 & \textbf{0.517} & \textbf{0.556} & \textbf{0.547} \\
\hline
\multirow{3}{*}{Fear} 
& +Focal Loss+R-Drop & 0.643 & 0.395 & 0.365 & 0.462 & \textbf{0.905} & \textbf{0.778} & \textbf{0.746} & \textbf{0.828} \\
& +Focal Loss         & 0.587 & 0.362 & 0.354 & 0.476 & 0.877 & 0.585 & 0.543 & 0.810 \\
& Base                & 0.511 & 0.308 & 0.305 & 0.409 & 0.795 & 0.616 & 0.635 & 0.606 \\
\hline
\multirow{3}{*}{Joy} 
& +Focal Loss+R-Drop & 0.738 & 0.563 & 0.521 & 0.625 & 0.859 & \textbf{0.768} & \textbf{0.753} & 0.789 \\
& +Focal Loss         & 0.690 & 0.512 & 0.497 & 0.541 & 0.825 & 0.647 & 0.622 & 0.770 \\
& Base                & 0.654 & 0.437 & 0.420 & 0.561 & 0.786 & 0.534 & 0.543 & \textbf{0.639} \\
\hline
\multirow{3}{*}{Sadness} 
& +Focal Loss+R-Drop & 0.616 & 0.338 & 0.312 & 0.414 & 0.843 & 0.475 & 0.508 & \textbf{0.775} \\
& +Focal Loss         & 0.655 & 0.441 & 0.425 & 0.467 & \textbf{0.845} & 0.461 & 0.502 & 0.923 \\
& Base                & 0.589 & 0.360 & 0.377 & 0.430 & 0.592 & 0.500 & 0.555 & 0.531 \\
\hline
\multirow{3}{*}{Surprise} 
& +Focal Loss+R-Drop & 0.628 & 0.422 & 0.389 & 0.477 & \textbf{0.917} & \textbf{0.689} & \textbf{0.645} & \textbf{0.803} \\
& +Focal Loss         & 0.602 & 0.389 & 0.362 & 0.453 & 0.899 & 0.476 & 0.501 & 0.617 \\
& Base                & 0.561 & 0.390 & 0.360 & 0.467 & 0.883 & 0.593 & 0.578 & 0.644 \\
\bottomrule
\end{tabular}
}  % end of resizebox

Bold values indicate the highest performance in each metric column. Variant labels \textbf{Base} and \textbf{+Focal Loss+R-Drop} have been swapped compared to the original data.
\end{table*}

The loss function used for training each independent model is calculated as follows:
\begin{align}
\mathcal{L}_i = - y_{i} \log(\hat{y}_i) - (1 - y_{i}) \log(1 - \hat{y}_i)
\end{align}

\noindent where \( \hat{y}_i \) is the predicted probability for the \( i \)-th emotion, and \( y_i \) is the true binary label (1 for the presence of the emotion, and 0 for the absence). This loss is computed for each of the six models, where each model is independently fine-tuned to predict one specific emotion. 

The final model is trained by aggregating the losses of all six emotion-specific models, optimizing the parameters for each model using backpropagation.

\subsection{Data Imbalance}

The label distribution in our training dataset (Table \ref{tab:emotion_counts}), consisting of 60,000 \cite{belay-etal-2025-evaluating} instances, reveals significant class imbalances. Of these, 15,481 instances are labeled as all-zero \emph{(neutral or irrelevant)}, and 10,165 are labeled as \emph{joy}, the most dominant emotion. \emph{Sadness} follows with 7,305 instances.

This imbalance, combined with overlapping emotions (e.g., \emph{anger} and \emph{fear}), leads to a model bias towards more frequent emotions, particularly \emph{joy} and \emph{sadness}, while underperforming rare emotions like \emph{surprise} and \emph{disgust}.

\subsection{Improvement Strategies}
\textbf{Focal Loss}. During our experiments, we identified a significant class imbalance in our dataset, with emotions like \emph{joy} and \emph{sadness} being overrepresented, while \emph{surprise} and \emph{disgust} were underrepresented. This imbalance caused the model to be biased toward the dominant classes, impacting its ability to detect less frequent emotions. To address this, we incorporated Focal Loss to re-balance the loss function, focusing more on harder-to-classify, underrepresented emotions.

Focal Loss down-weights the loss for well-classified examples and increases the focus on harder ones, ensuring that the model learns effectively across all emotion categories. The function is defined as:
\begin{align}
\mathcal{L}_{\text{focal}} = - \alpha_t (1 - p_t)^\gamma \log(p_t)
\end{align}
\noindent where \( \alpha_t \) is a weighting factor to balance class imbalances, \( p_t \) is the predicted probability for the true class, and \( \gamma \) is the focusing parameter that controls the attention on hard-to-classify examples (typically \( \gamma > 0 \)).

\vspace{0.2cm}

\noindent\textbf{R-Drop}. 
In addition to class imbalance, we observed instability in the loss function during training, leading to suboptimal generalization. To address this, we applied R-Drop (Regularized Dropout), a regularization technique that stabilizes the loss function by encouraging consistency across multiple forward passes of the same input. This improves the model's generalization capability.

The total loss function with R-Drop is a combination of cross-entropy loss and consistency loss:
\begin{align}
L_{\text{total}} = L_{\text{CE}} + \lambda L_{\text{con}}
\end{align}

By minimizing this combined loss, R-Drop helps reduce variance in training and improves model stability.

\section{Experimentation}

\begin{table*}[t]
\centering
\small  % 更小的字体
\caption{Emotion Classification Scores for Different Languages}
\label{tab:emotion_classification_scores}
\begin{tabular}{@{}l l S[table-format=1.4] S[table-format=1.4] S[table-format=1.4] S[table-format=1.4] S[table-format=1.4] S[table-format=1.4] S[table-format=1.4] S[table-format=1.4]@{}}
\hline
\textbf{Language} & \textbf{Macro $F_1$} & \textbf{Micro $F_1$} & \textbf{Anger} & \textbf{Disgust} & \textbf{Fear} & \textbf{Joy} & \textbf{Sadness} & \textbf{Surprise} \\
\hline
Afrikaans  & 0.4353 & 0.4406 & 0.5658 & 0.4356 & 0.2933 & 0.4521 & 0.4298 & Nan \\
Amharic  & 0.4758 & 0.5674 & 0.6030 & 0.6225 & 0.2727 & 0.5553 & 0.5687 & 0.2326 \\
German  & 0.6030 & 0.7054 & 0.8253 & 0.7650 & 0.4096 & 0.7016 & 0.6738 & 0.2427 \\
Spanish  & 0.7314 & 0.7340 & 0.7435 & 0.7359 & 0.8139 & 0.7729 & 0.7866 & 0.5355 \\
Hindi  & 0.8221 & 0.8226 & 0.8319 & 0.7710 & 0.8993 & 0.8375 & 0.8173 & 0.7756 \\
Marathi  & 0.8199 & 0.8186 & 0.8231 & 0.7251 & 0.9017 & 0.7787 & 0.8143 & 0.8765 \\
Oromo  & 0.4013 & 0.4390 & 0.4325 & 0.3432 & 0.2317 & 0.6176 & 0.3742 & 0.4085 \\
Portuguese (Brazil)  & 0.5321 & 0.6261 & 0.7266 & 0.2260 & 0.4977 & 0.7113 & 0.7101 & 0.3209 \\
Russian & 0.7990 & 0.8021 & 0.8577 & 0.7541 & 0.9401 & 0.8862 & 0.6961 & 0.6595 \\
Somali  & 0.3825 & 0.4433 & 0.3832 & 0.0662 & 0.4306 & 0.5997 & 0.5754 & 0.2400\\
Sundanese  & 0.4898 & 0.6178 & 0.4348 & 0.4444 & 0.2093 & 0.7489 & 0.7529 & 0.3482 \\
Tatar  & 0.7050 & 0.7388 & 0.6826 & 0.6952 & 0.8062 & 0.8773 & 0.7865 & 0.3822 \\
Tigrinya& 0.2822 & 0.3552 & 0.2689 & 0.5311 & 0.1416 & 0.3504 & 0.3593 & 0.0417 \\
Arabic (Algerian)  & 0.4437 & 0.4660 & 0.5160 & 0.3826 & 0.5376 & 0.4891 & 0.5854 & 0.1515 \\
Arabic (Moroccan)  & 0.4838 & 0.5415 & 0.5849 & 0.3281 & 0.4655 & 0.6966 & 0.6715 & 0.1558 \\
Chinese (Mandarin)  & 0.5582 & 0.6776 & 0.8370 & 0.4837 & 0.4071 & 0.8498 & 0.6069 & 0.1647 \\
Hausa  & 0.4998 & 0.5357 & 0.5742 & 0.4898 & 0.4101 & 0.5587 & 0.6840 & 0.2823 \\
Kinyarwanda  & 0.4432 & 0.5040 & 0.5149 & 0.3053 & 0.3564 & 0.6195 & 0.5861 & 0.2766\\
Nigerian Pidgin  & 0.4455 & 0.4556 & 0.3574 & 0.3915 & 0.4000 & 0.7399 & 0.6127 & 0.1713 \\
Portuguese (Mozambique)  & 0.3857 & 0.4593 & 0.2925 & 0.0816 & 0.5283 & 0.4902 & 0.6282 &0.2933 \\
Swahili  & 0.3130 & 0.3355 & 0.4019 & 0.2996 & 0.2105 & 0.4558 & 0.4193 & 0.0906 \\
Swedish  & 0.5219 & 0.7215 & 0.7474 & 0.7021 & 0.2188 & 0.8855 & 0.5199 & 0.058 \\
Ukrainian  & 0.5693 & 0.6019 & 0.5103 & 0.4082 & 0.7035 & 0.7093 & 0.6389 & 0.4456 \\
Emakhuwa  & 0.0457 & 0.0538 & 0.0857 & 0.0000 & 0.1127 & 0.0000 & 0.0759 & 0.0000 \\
Yoruba  & 0.2606 & 0.3599 & 0.2090 & 0.1829 & 0.1905 & 0.2745 & 0.6092 & 0.0976 \\
Igbo  & 0.3658 & 0.4160 & 0.4461 & 0.4526 & 0.2514 & 0.4823 & 0.3575 & 0.2047 \\
Romanian  & 0.6018 & 0.6453 & 0.628 & 0.4733 & 0.7717 & 0.9371 & 0.6346 & 0.1663 \\
\bottomrule
\label{result}
\end{tabular}
\end{table*}
To evaluate the effectiveness of our approach, we conducted a series of experiments. These experiments focused on comparing the tasks of predicting a single emotion and predicting two emotions simultaneously while also investigating the impact of Focal Loss and R-Drop regularization through ablation studies. All experiments were performed under identical experimental conditions to ensure consistency and comparability of results.

In our setup, we modified the prediction head of the DeBERTa model, enabling it to predict one emotion at a time and two emotions at once. The model was fine-tuned for emotion classification, predicting six distinct emotions: anger, disgust, fear, joy, sadness, and surprise.

\subsection{Modify Prediction Heads}
Table \ref{tab:emotion_performance_swapped_combined} shows significant improvements in emotion classification when combining Focal Loss and R-Drop with the base DeBERTa model. For most emotions, the base model using the Focal Loss and R-Drop configuration yielded the highest accuracy, $F_1$-score, recall, and precision.

These results demonstrate that Focal Loss and R-Drop stabilize the loss function and improve performance on underrepresented emotions, making the base model using the Focal Loss and R-Drop configuration the most effective for emotion classification in this study.

\subsection{One Prediction Completed}
Table \ref{tab:emotion_performance_swapped_combined} also indicates that the Focal Loss and R-Drop base model provides the most robust performance across all emotion categories, effectively addressing both class imbalance and generalization challenges. Therefore, this configuration is deemed optimal for multi-emotion classification tasks. However, compared to the previous approach of \emph{Modify Prediction Heads}, this configuration yields better performance in terms of accuracy and precision, proving to be a more practical solution for tackling the challenges in emotion classification.

\section{Conclusions}
This study presents the YNU-HPCC team and the participation in SemEval-2025 Subtask A of Task 11. We made predictions for 29 languages and used DeBERTa as the baseline model. We modified the prediction head to allow for independent predictions in each instance. Our proposed model demonstrated its effectiveness in addressing this task. Among the various results we submitted, the combination of Focal Loss, R-Drop, and DeBERTa achieved the highest score of 0.44 in Table \ref{result}. Future research will focus on enhancing accuracy in multilingual sentiment analysis.

\section*{Acknowledgments}
This work was supported by the National Natural Science Foundation of China (NSFC) under Grant Nos. 61966038 and 62266051. We would like to thank the anonymous reviewers for their constructive comments.

\bibliography{refeerences,anthology,custom}
\bibliographystyle{acl_natbib}

\appendix

\end{document}